\documentclass[letterpaper, 10 pt, conference]{ieeeconf}  

\usepackage[top=54pt, bottom=54pt, left=0.75in, right=0.75in]{geometry}

\IEEEoverridecommandlockouts                              

\usepackage{tabularx}
\usepackage{array}

\usepackage{float} 

\usepackage{amssymb}
\usepackage{newtxtext,newtxmath} 
\usepackage{graphicx}
\usepackage[export]{adjustbox}
\usepackage{url}
\usepackage{latexsym}
\usepackage{algpseudocode}
\usepackage{amsmath}
\usepackage{lipsum}
\usepackage{caption}
\usepackage{bbding}
\usepackage{wasysym}
\usepackage{blindtext}
\usepackage{booktabs}
\usepackage{tabularx}
\usepackage{array}
\usepackage{pifont}
\usepackage[export]{adjustbox}
\usepackage{float}
\usepackage{multirow}

\usepackage{amsthm}
\usepackage{xcolor}
\usepackage{cite}

\usepackage{subcaption}

\title{\LARGE \bf
Integrating Vision Foundation Models with Reinforcement Learning for Enhanced Object Interaction
}

\author{Ahmad Farooq*$^{1}$ and Kamran Iqbal$^{2}$
\thanks{This work was not supported by any organization.}
\thanks{*Corresponding Author: Ahmad Farooq}
\thanks{$^{1}$Ahmad Farooq is a Ph.D. Candidiate in the Electrical and Computer Engineering Department at the University of Arkansas at Little Rock, AR, 72204, USA
        {\tt\small afarooq@ualr.edu}}%
\thanks{$^{2}$Kamran Iqbal is a Professor in the Electrical and Computer Engineering Department at the University of Arkansas at Little Rock, AR, 72204, USA
        {\tt\small kxiqbal@ualr.edu}}%
\thanks{This is the author's preprint version of an article published in RCVE'25: Proceedings of the 2025 3rd International Conference on Robotics, Control and Vision Engineering. The final published version is available in the ACM Digital Library: https://doi.org/10.1145/3747393.3747399.}%
}

\begin{document}

\maketitle

\thispagestyle{empty}
\pagestyle{empty}

\begin{abstract}
This paper presents a novel approach that integrates vision foundation models with reinforcement learning to enhance object interaction capabilities in simulated environments. By combining the Segment Anything Model (SAM) and YOLOv5 with a Proximal Policy Optimization (PPO) agent operating in the AI2-THOR simulation environment, we enable the agent to perceive and interact with objects more effectively. Our comprehensive experiments, conducted across four diverse indoor kitchen settings, demonstrate significant improvements in object interaction success rates and navigation efficiency compared to a baseline agent without advanced perception. The results show a 68\% increase in average cumulative reward, a 52.5\% improvement in object interaction success rate, and a 33\% increase in navigation efficiency. These findings highlight the potential of integrating foundation models with reinforcement learning for complex robotic tasks, paving the way for more sophisticated and capable autonomous agents.


\end{abstract}

\section{Introduction}

\begin{figure*}[htpb]
\centering
\begin{subfigure}[b]{0.24\textwidth}
    \includegraphics[width=\textwidth]{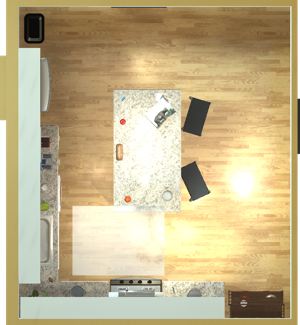}
    \caption{FloorPlan1}
    \label{fig:floorplan1}
\end{subfigure}
\hfill
\begin{subfigure}[b]{0.24\textwidth}
    \includegraphics[width=\textwidth]{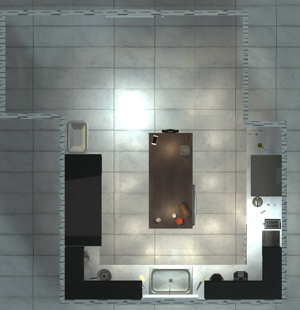}
    \caption{FloorPlan2}
    \label{fig:floorplan2}
\end{subfigure}
\hfill
\begin{subfigure}[b]{0.24\textwidth}
    \includegraphics[width=\textwidth]{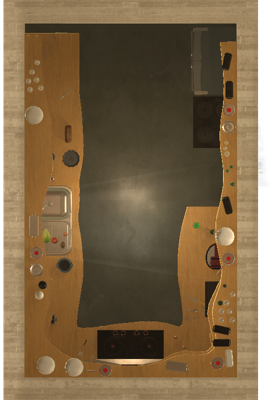}
    \caption{FloorPlan3}
    \label{fig:floorplan3}
\end{subfigure}
\hfill
\begin{subfigure}[b]{0.24\textwidth}
    \includegraphics[width=\textwidth]{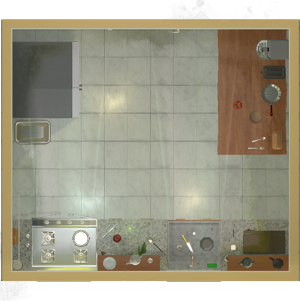}
    \caption{FloorPlan4}
    \label{fig:floorplan4}
\end{subfigure}
\caption{Top-down views of the four kitchen environments in AI2-THOR used in our experiments. Each environment presents distinct challenges due to differences in layout and object placement.}
\label{fig:kitchen_floorplans}
\end{figure*}

The development of autonomous agents capable of interacting with complex environments is a fundamental goal in robotics and artificial intelligence. A key challenge in achieving this goal lies in equipping agents with advanced perception and decision-making abilities that enable them to understand and manipulate their surroundings effectively. Robust perception allows agents to recognize and localize objects, comprehend spatial relationships, and interpret dynamic scenes \cite{he2017mask,redmon2017yolo9000}. Reinforcement Learning (RL) provides a framework for agents to learn optimal behaviors through trial-and-error interactions with the environment \cite{sutton2018reinforcement,schulman2017proximal}.

Recent advancements in computer vision and RL have led to significant progress in tasks such as visual navigation \cite{zhu2017target,gupta2017cognitive}, object manipulation \cite{kalashnikov2018qt,andrychowicz2020learning}, and human-robot interaction \cite{shao2020concept2robot,tellex2020robots}. However, integrating sophisticated perception models with RL agents remains a challenging problem due to factors such as the complexity of real-time processing in dynamic environments and computational overhead.

Vision foundation models, including YOLOv5 \cite{jocher2020yolov5} and Segment Anything Model (SAM) \cite{kirillov2023segment}, have demonstrated exceptional capabilities in object detection and segmentation tasks. These models are pre-trained on extensive datasets and can generalize across various domains, reducing the dependence on task-specific training data. By integrating these models with RL agents, we can enhance the agents' perceptual understanding, leading to improved performance in tasks that involve complex interactions with the environment.

In this work, we propose a novel approach that combines vision foundation models with reinforcement learning to enhance object interaction capabilities in simulated environments. Specifically, we integrate SAM and YOLOv5 into the perception pipeline of a Proximal Policy Optimization (PPO) agent operating within the AI2-THOR simulation environment \cite{kolve2017ai2}. The AI2-THOR environment offers richly interactive 3D scenes, providing a suitable platform for training agents in object interaction and navigation tasks.

Our approach addresses several key challenges:

\begin{itemize}
    \item \textbf{Perception Integration}: We develop a method to effectively incorporate the outputs of SAM and YOLOv5 into the agent's observation space, allowing for enhanced scene understanding without incurring prohibitive computational costs.
    \item \textbf{Reward Function Design}: We formulate a reward function that balances exploration, object interaction, and goal achievement, guiding the agent to learn efficient strategies for interacting with objects.
    \item \textbf{Policy Learning}: We design a policy network architecture that leverages the enriched perceptual input to learn effective policies for object interaction tasks.
\end{itemize}

Our main contributions are as follows:

\begin{enumerate}
    \item We present a novel framework that integrates vision foundation models with reinforcement learning agents to enhance object interaction capabilities.
    \item We develop techniques to efficiently incorporate advanced perception outputs into the RL framework, addressing computational and integration challenges.
    \item We conduct extensive experiments demonstrating significant performance improvements over a baseline agent, highlighting the effectiveness of our approach.
    \item We provide insights into the implications of integrating foundation models with reinforcement learning, suggesting avenues for future research.
\end{enumerate}

The rest of the paper is organized as follows. In Section~\ref{sec:related_work}, we review related work on reinforcement learning in robotics and the integration of perception models. Section~\ref{sec:methodology} details our proposed approach, including the integration of SAM and YOLOv5 with the RL agent and the design of the reward function. Section~\ref{sec:experiments} describes our experimental setup and evaluation metrics. We present and analyze the results in Section~\ref{sec:results}, and discuss the implications of our findings in Section~\ref{sec:discussion}. We conclude the paper in Section~\ref{sec:conclusion}, and outline directions for future research in Section~\ref{sec:future_work}.

\section{Related Work}
\label{sec:related_work}

In this section, we review the literature on reinforcement learning in robotics, the integration of advanced perception models with RL agents, vision foundation models, and the use of simulation environments for training and evaluation.

\subsection{Reinforcement Learning in Robotics}

Reinforcement learning has been extensively applied to robotic control tasks, enabling agents to learn complex behaviors through interactions with the environment \cite{kober2013reinforcement,sutton2018reinforcement}. Model-free RL algorithms, such as Proximal Policy Optimization (PPO) \cite{schulman2017proximal} and Deep Deterministic Policy Gradients (DDPG) \cite{lillicrap2015continuous}, have shown success in continuous control problems. Applications include robotic manipulation \cite{levine2016end,kalashnikov2018qt}, locomotion \cite{schulman2015trust,heess2017emergence}, and navigation \cite{zhu2017target,gupta2017cognitive}.

Challenges in applying RL to real-world robotics include sample inefficiency, safety during exploration, and the sim-to-real gap \cite{zhao2020sim,zhang2018deep}. Addressing these issues often involves leveraging simulation environments and incorporating prior knowledge or advanced perception.

\subsection{Integration of Perception Models with RL}

Integrating perception models with RL agents enhances their ability to interpret and interact with the environment. Early works utilized convolutional neural networks (CNNs) to process raw images for end-to-end learning \cite{mnih2015human}. Subsequent research incorporated object detection and semantic segmentation to provide richer observations \cite{mousavian2019visual,chaplot2020object}.

Recent studies have explored combining RL with advanced vision models to improve scene understanding. For instance, Zhu et al.  \cite{zhu2017target} developed a target-driven visual navigation framework that enables agents to find specific targets in indoor environments using deep RL without requiring feature engineering or 3D reconstruction. Wang et al.  \cite{wang2019reinforced} proposed a Reinforced Cross-Modal Matching approach that uses both extrinsic environmental rewards and intrinsic cycle-reconstruction rewards to better align navigation trajectories with natural language instructions. 

However, the integration of large-scale foundation models like SAM and YOLOv5 into RL agents is still an emerging area, with potential to significantly boost performance in complex tasks.

\subsection{Vision Foundation Models}

Vision foundation models are pre-trained on large datasets and can generalize across various tasks. YOLOv5 \cite{jocher2020yolov5} is renowned for real-time object detection with high accuracy, making it suitable for applications requiring quick responses. The Segment Anything Model (SAM) \cite{kirillov2023segment} provides prompt-based segmentation without additional training, enabling zero-shot generalization to new objects and scenes.

The deployment of these models in robotics has been facilitated by advances in computational hardware and optimization techniques \cite{ren2015faster,he2017mask}. Integrating such models into RL agents offers the promise of enhanced perception without the need for extensive task-specific data collection and training.

\subsection{Simulation Environments}

Simulation environments like AI2-THOR \cite{kolve2017ai2}, Habitat \cite{savva2019habitat}, and Gibson \cite{xiazamirhe2018gibsonenv} provide realistic and interactive platforms for training RL agents. These environments enable safe and efficient exploration, reducing the risks and costs associated with real-world experimentation.

AI2-THOR offers high-fidelity 3D scenes with physics-based interactions, making it suitable for tasks involving object manipulation and navigation. Our work leverages AI2-THOR to integrate advanced perception models with RL agents, providing a platform to evaluate the effectiveness of our approach in complex, interactive settings.

\section{Methodology}
\label{sec:methodology}

In this section, we describe our approach to integrating vision foundation models with a reinforcement learning agent to enhance object interaction capabilities. We outline the framework for the RL agent, the perception pipeline, the reward function design, and the policy network architecture.

\begin{figure*}[htbp]
\centering
\includegraphics[width=0.9\textwidth]{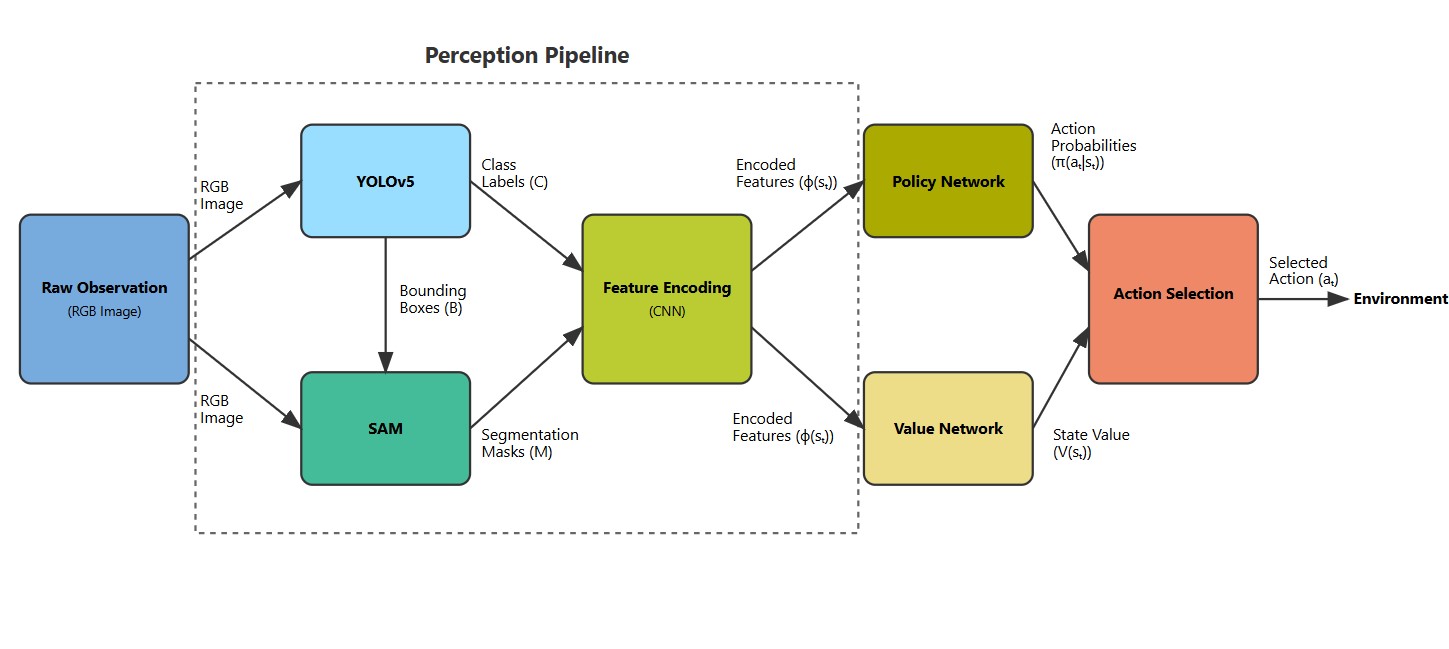}
\caption{System architecture integrating SAM and YOLOv5 with the RL agent. The perception pipeline processes the raw observation to extract rich features, which are then used by the policy network to make action decisions.}
\label{fig:system_architecture}
\end{figure*}

Our goal is to enable an RL agent to effectively perceive and interact with objects in a complex environment by leveraging the advanced capabilities of SAM and YOLOv5. Figure~\ref{fig:system_architecture} illustrates the overall system architecture.

\subsection{Framework for RL Agent}

We utilize the AI2-THOR simulation environment \cite{kolve2017ai2}, focusing on four kitchen scenes: FloorPlan1 to FloorPlan4. Each scene presents unique spatial configurations and object placements, providing diverse challenges for the agent.

The agent's action space $\mathcal{A}$ includes:

\begin{itemize}
    \item \textbf{Navigation Actions}: \textit{MoveAhead}, \textit{RotateLeft}, \textit{RotateRight}, \textit{LookUp}, \textit{LookDown}.
    \item \textbf{Interaction Actions}: \textit{PickupObject}, \textit{DropObject}.
\end{itemize}

The observation space $\mathcal{O}$ consists of RGB images captured from the agent's first-person perspective at each time step $t$.

\subsection{Perception Pipeline}

To enhance the agent's perception, we integrate SAM \cite{kirillov2023segment} and YOLOv5 \cite{jocher2020yolov5} into the observation processing pipeline.

\subsubsection{YOLOv5 for Object Detection}

YOLOv5 processes the raw RGB image to detect objects, outputting bounding boxes $B = \{b_i\}_{i=1}^N$ and class labels $C = \{c_i\}_{i=1}^N$, where $N$ is the number of detected objects. This provides the agent with information about the presence and location of objects.

\subsubsection{SAM for Object Segmentation}

Using the bounding boxes from YOLOv5 as prompts, SAM generates segmentation masks $M = \{m_i\}_{i=1}^N$, providing precise object contours and spatial relationships. This detailed segmentation aids the agent in understanding the environment's structure.

\subsubsection{Feature Encoding}

The combined outputs $(B, C, M)$ are encoded using a convolutional neural network (CNN) to produce a feature representation $\phi(s_t)$, where $s_t$ is the state at time $t$. This representation captures both the semantic and spatial information necessary for decision-making.

\subsection{Reward Function Design}

Designing an effective reward function is critical for guiding the agent's learning. We define the reward $r_t$ at time $t$ as:

\begin{equation}
    r_t = \alpha \cdot \Delta d_t + \beta \cdot s_t - \gamma \cdot c_t
\end{equation}

where:

\begin{itemize}
    \item $\Delta d_t = d_{t-1} - d_t$ is the change in distance to the target object, encouraging the agent to approach the target.
    \item $s_t$ is a success indicator, $s_t = 1$ if the agent successfully interacts with the target object, and $0$ otherwise.
    \item $c_t$ is a penalty for collisions or invalid actions.
    \item $\alpha$, $\beta$, and $\gamma$ are weighting factors tuned empirically.
\end{itemize}

This reward structure incentivizes efficient navigation, successful interaction, and discourages undesirable behaviors.

\subsection{Policy Network Architecture}

The policy network consists of two main components:

\begin{enumerate}
    \item \textbf{Perception Encoder}: A CNN that processes the feature representation $\phi(s_t)$ to extract high-level features.
    \item \textbf{Policy and Value Heads}: Two fully connected layers that output the action probabilities $\pi(a_t|s_t)$ and state value estimates $V(s_t)$.
\end{enumerate}

We employ the PPO algorithm \cite{schulman2017proximal} to optimize the policy, utilizing a clipped surrogate objective to ensure stable learning.

Integrating large perception models introduces computational overhead. To address this, we optimize the inference pipelines and utilize batch processing where possible. We also design an efficient feature encoding strategy that captures essential information without overcomplicating the state representation.

\section{Experiments}
\label{sec:experiments}

We conduct a series of experiments to evaluate the effectiveness of our approach. This section details the experimental setup, baseline comparisons, implementation details, and evaluation metrics.

\subsection{Experimental Setup}

Our experiments are carried out in the AI2-THOR environment \cite{kolve2017ai2}, focusing on the four kitchen scenes (FloorPlan1--FloorPlan4) depicted in Figure~\ref{fig:kitchen_floorplans}. Each scene provides a unique set of challenges due to variations in layout, object types, and spatial arrangements.

The agent is tasked with navigating to and interacting with a target object specified at the beginning of each episode. The target objects vary and include items like \textit{Mug}, \textit{Apple}, \textit{Knife}, and so on. The agent must locate the object, navigate to it, and perform an appropriate interaction (e.g., \textit{PickupObject}).

\subsection{Baseline}

We compare our perception-enhanced agent against a Standard RL Agent using raw RGB observations without advanced perception models, trained with the same RL algorithm and hyperparameters. The baseline allows us to assess the benefits of integrating advanced perception models over traditional observation modalities.

\subsection{Training Procedure}

The agent is trained by interacting with the environment, collecting trajectories $\tau = \{(s_t, a_t, r_t, s_{t+1})\}$, and updating the policy network using gradient ascent on the PPO objective. The perception models (SAM and YOLOv5) are kept fixed during training to leverage their pre-trained capabilities without incurring the computational cost of fine-tuning.

\subsection{Implementation Details}

All agents are implemented using PyTorch \cite{paszke2019pytorch}. The perception-enhanced agent utilizes pre-trained SAM and YOLOv5 models, with inference optimized for real-time performance. We ensure consistent training conditions across all agents, including the same random seeds, to facilitate fair comparisons.

The hyperparameters are set as follows:

\begin{itemize}
    \item \textbf{Learning Rate}: $3 \times 10^{-4}$
    \item \textbf{Discount Factor ($\gamma$)}: 0.99
    \item \textbf{GAE Parameter ($\lambda$)}: 0.95
    \item \textbf{PPO Clip Parameter ($\epsilon$)}: 0.2
    \item \textbf{Batch Size}: 64
    \item \textbf{Epochs per Update}: 4
\end{itemize}

Experiments are conducted on a workstation with an NVIDIA RTX 3080 GPU and 64 GB of RAM. Resource utilization is monitored to assess the computational overhead introduced by the perception models.

Agents are trained for $1 \times 10^6$ time steps, with performance evaluated at regular intervals. We use early stopping if the performance plateaus. Training curves are recorded for analysis of learning dynamics.

During training and evaluation, we collect detailed logs of the agents' actions, observations, rewards, and internal states. This data supports quantitative analysis and aids in diagnosing any performance issues.

To ensure the robustness of our results, we repeat each experiment five times with different random seeds and report the mean and standard deviation of the metrics.

\subsection{Evaluation Metrics}

We evaluate the agents using the following metrics:

\begin{itemize}
    \item \textbf{Success Rate (\%)}: The percentage of episodes in which the agent successfully interacts with the target object.
    \item \textbf{Average Cumulative Reward}: The mean total reward accumulated per episode.
    \item \textbf{Navigation Efficiency (\%)}: The ratio of the optimal path length to the actual path length taken by the agent, averaged over successful episodes, multiplied by 100.
    \item \textbf{Interaction Efficiency}: The number of interaction attempts made before a successful interaction, measuring the agent's ability to correctly perform tasks without unnecessary actions.
\end{itemize}

\begin{figure*}[htbp]
\centering
\begin{subfigure}[b]{0.24\textwidth}
    \includegraphics[width=\textwidth]{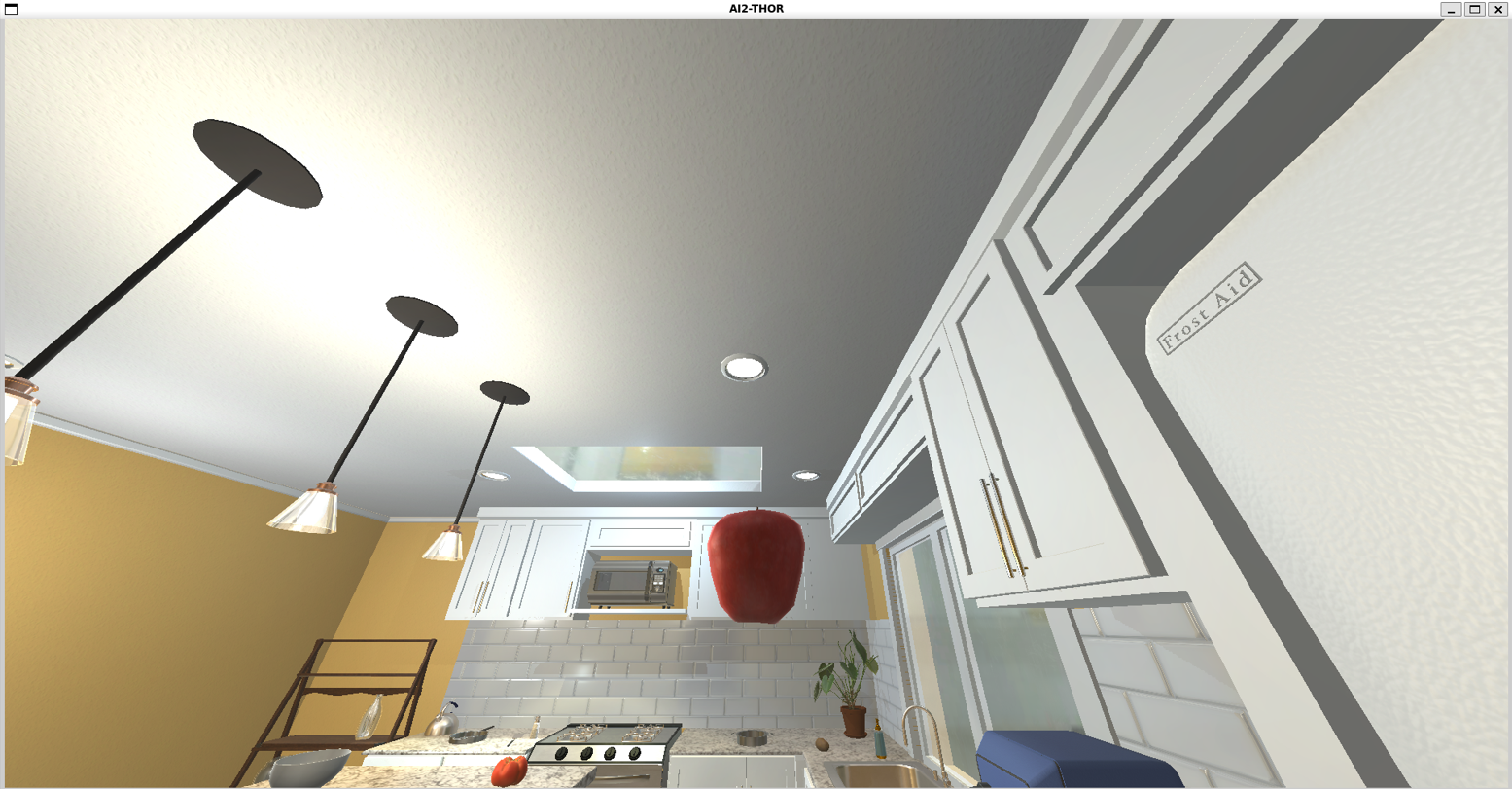}
    \caption{An apple in FloorPlan1}
    \label{fig:apple}
\end{subfigure}
\hfill
\begin{subfigure}[b]{0.24\textwidth}
    \includegraphics[width=\textwidth]{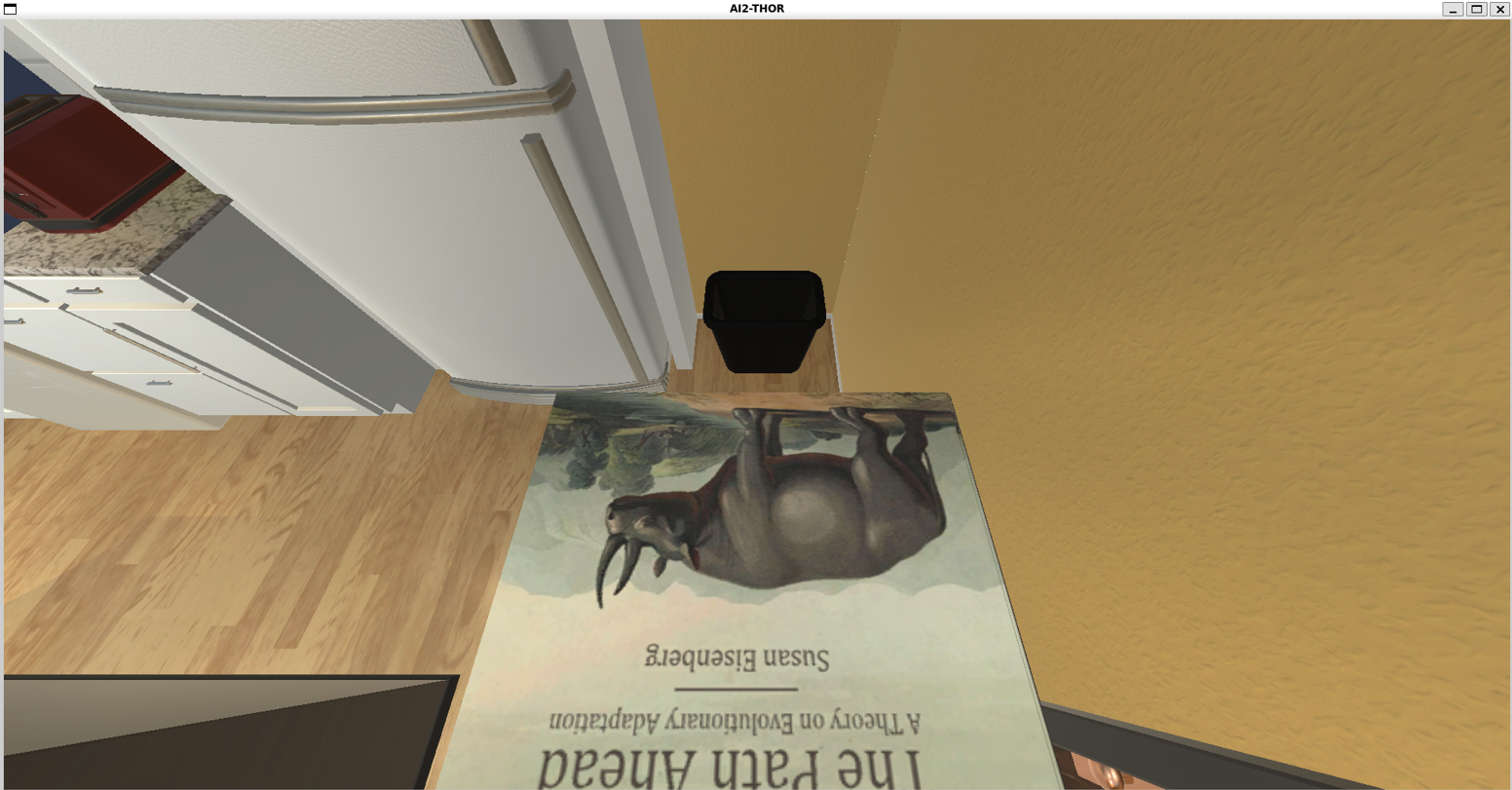}
    \caption{A book in FloorPlan1}
    \label{fig:book}
\end{subfigure}
\hfill
\begin{subfigure}[b]{0.24\textwidth}
    \includegraphics[width=\textwidth]{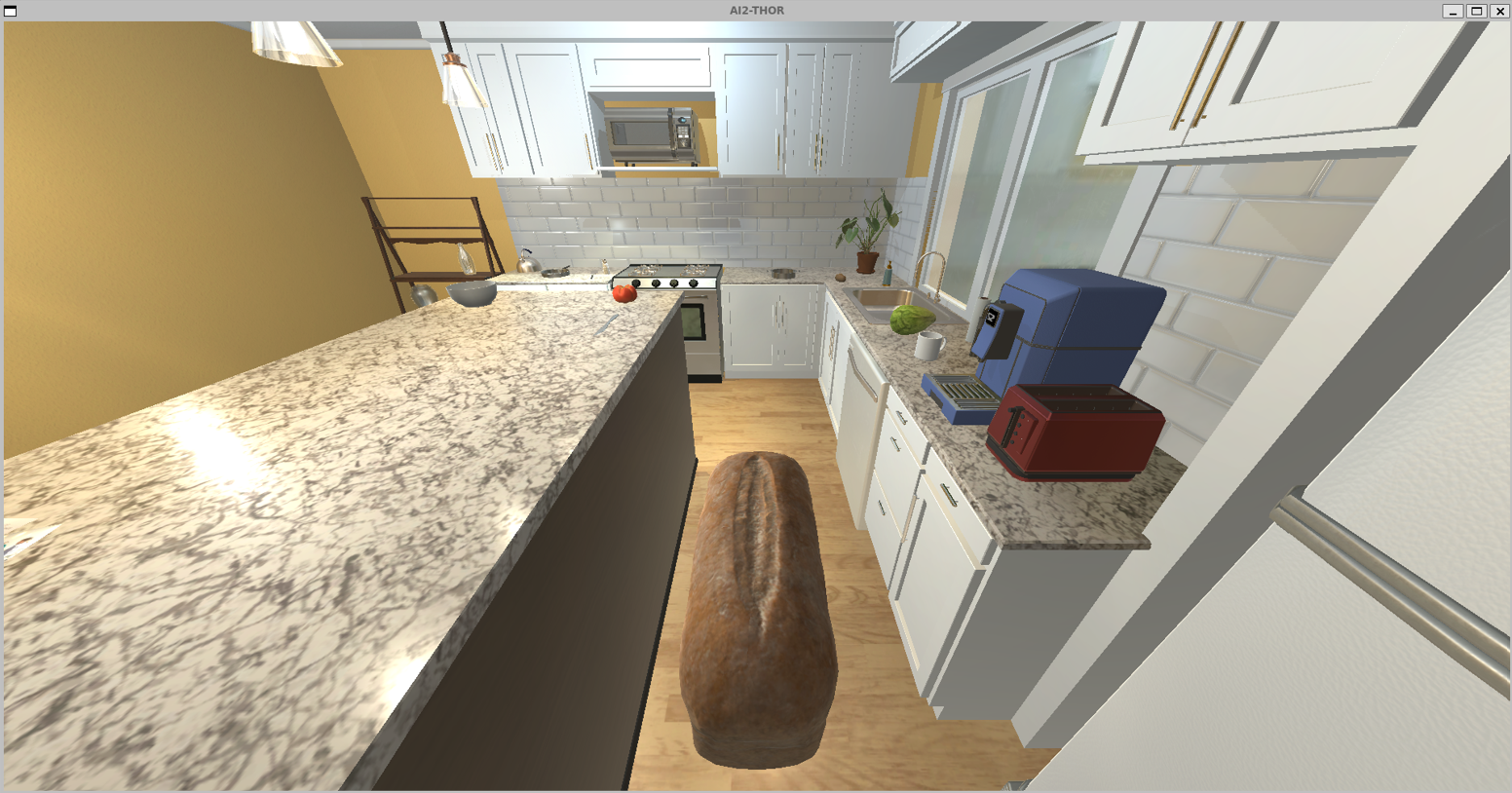}
    \caption{A bread in FloorPlan1}
    \label{fig:bread}
\end{subfigure}
\hfill
\begin{subfigure}[b]{0.24\textwidth}
    \includegraphics[width=\textwidth]{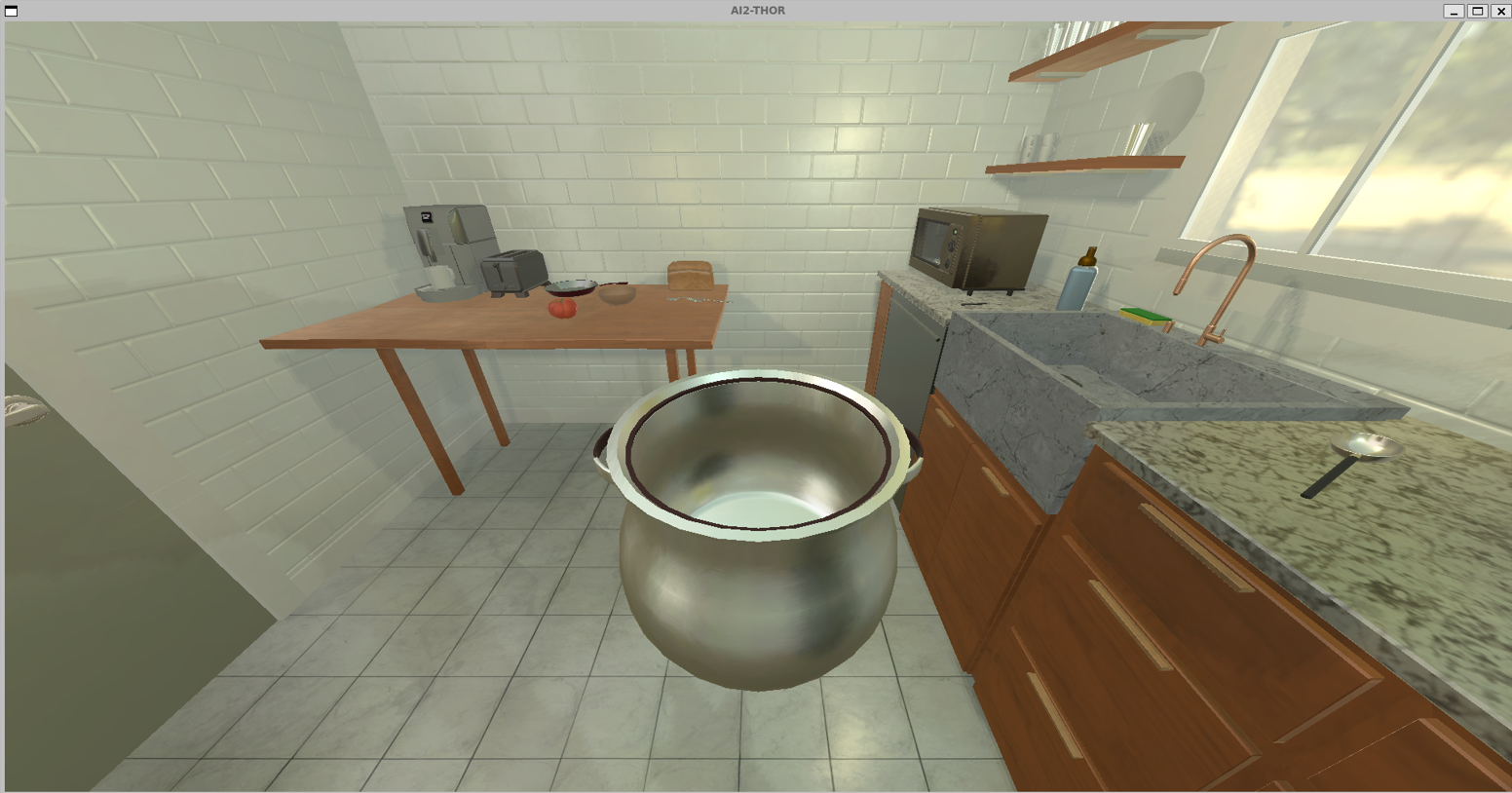}
    \caption{A cooking pot in FloorPlan4}
    \label{fig:cooking_pot}
\end{subfigure}

\vspace{0.3cm} 

\begin{subfigure}[b]{0.24\textwidth}
    \includegraphics[width=\textwidth]{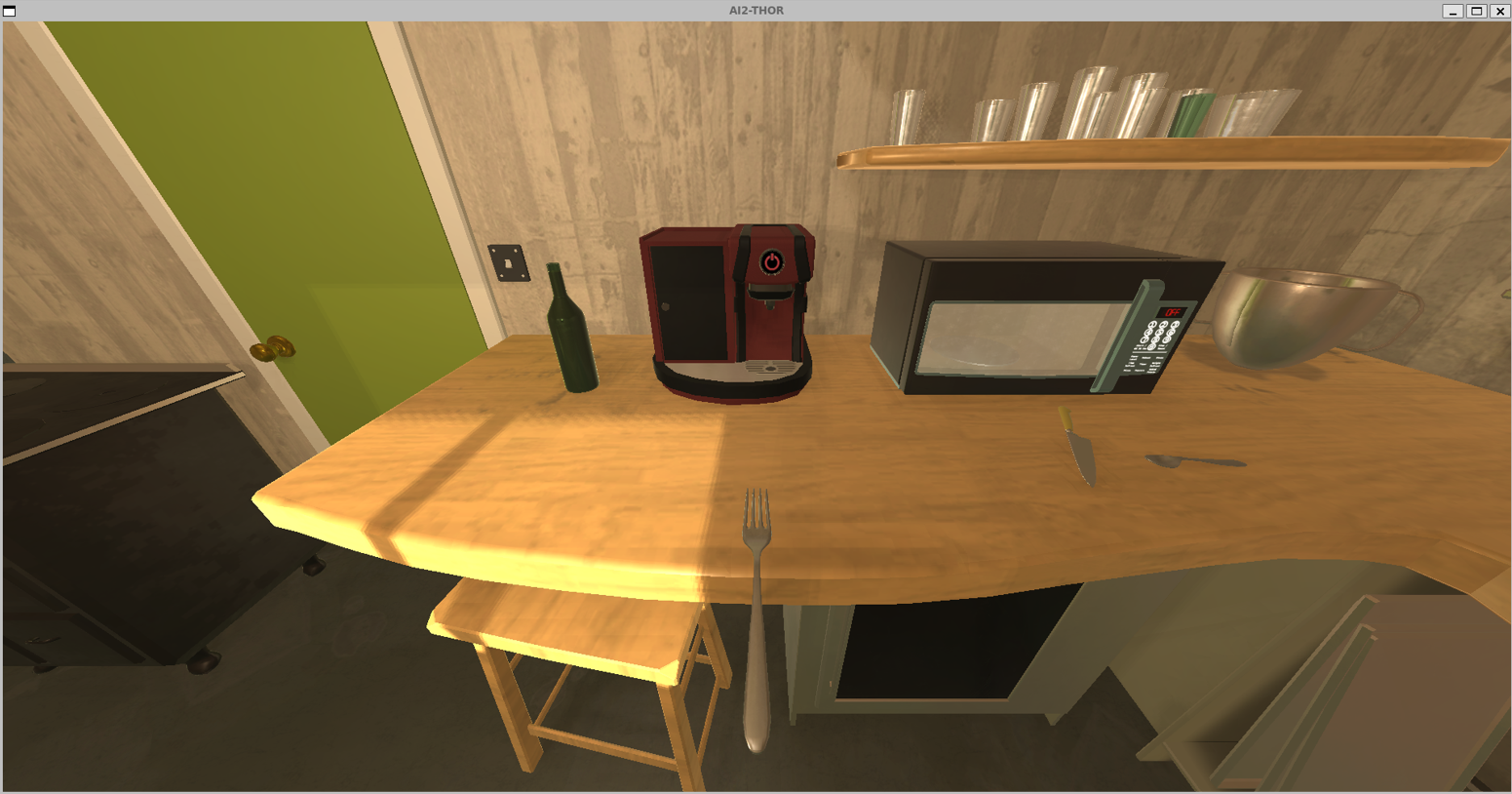}
    \caption{A fork in FloorPlan3}
    \label{fig:fork}
\end{subfigure}
\hfill
\begin{subfigure}[b]{0.24\textwidth}
    \includegraphics[width=\textwidth]{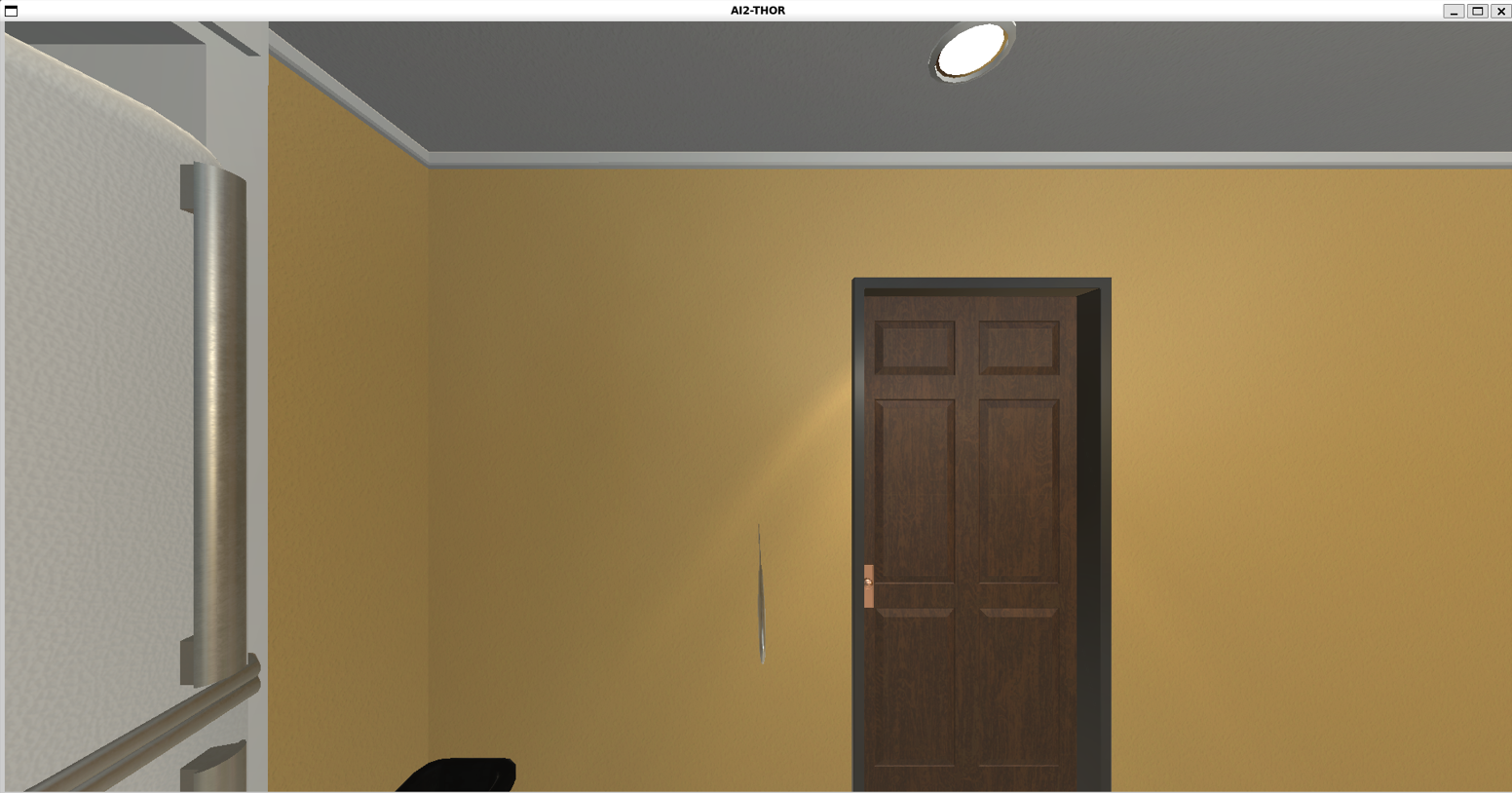}
    \caption{A knife in FloorPlan1}
    \label{fig:knife}
\end{subfigure}
\hfill
\begin{subfigure}[b]{0.24\textwidth}
    \includegraphics[width=\textwidth]{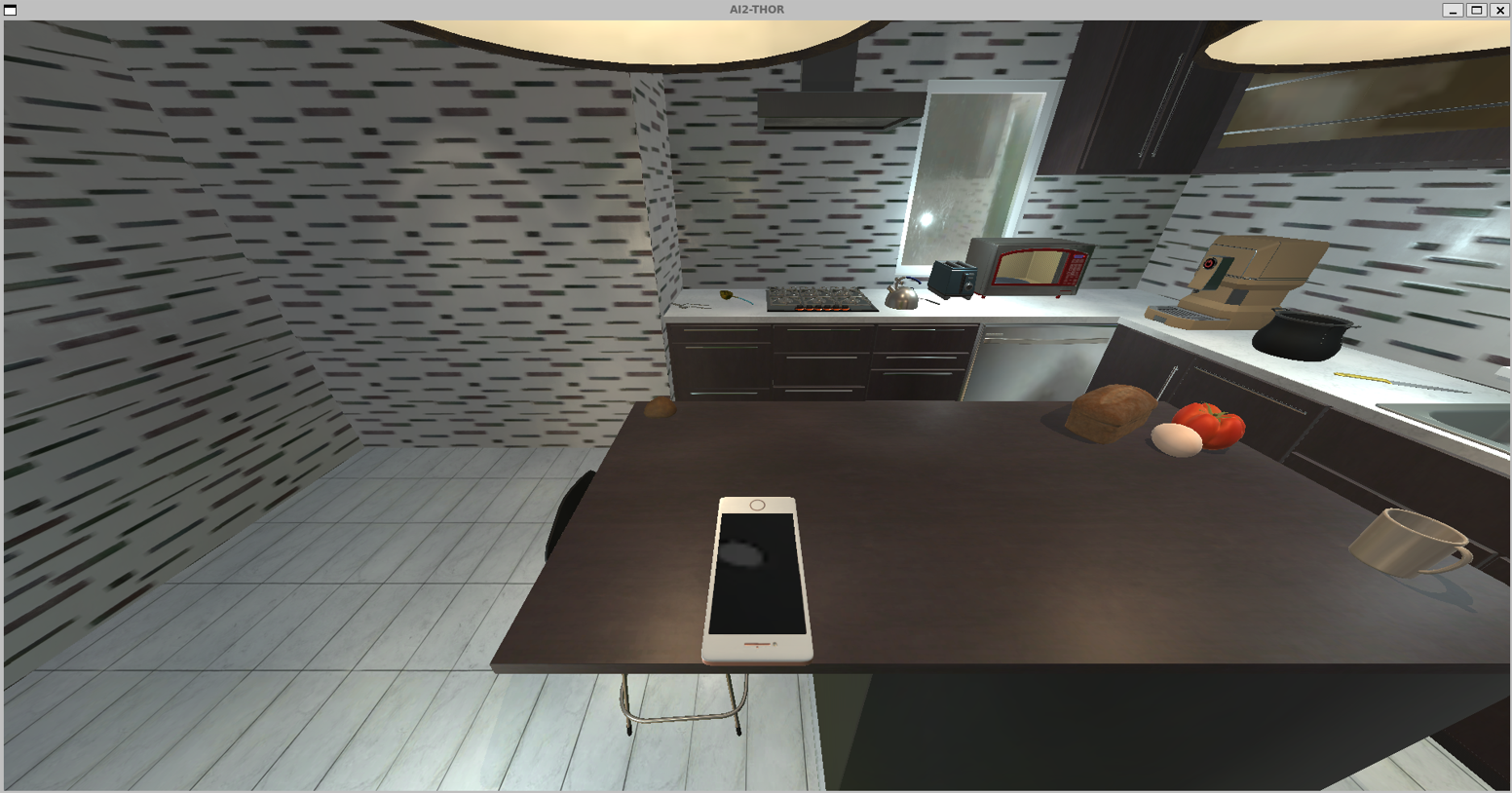}
    \caption{A phone in FloorPlan2}
    \label{fig:phone}
\end{subfigure}
\hfill
\begin{subfigure}[b]{0.24\textwidth}
    \includegraphics[width=\textwidth]{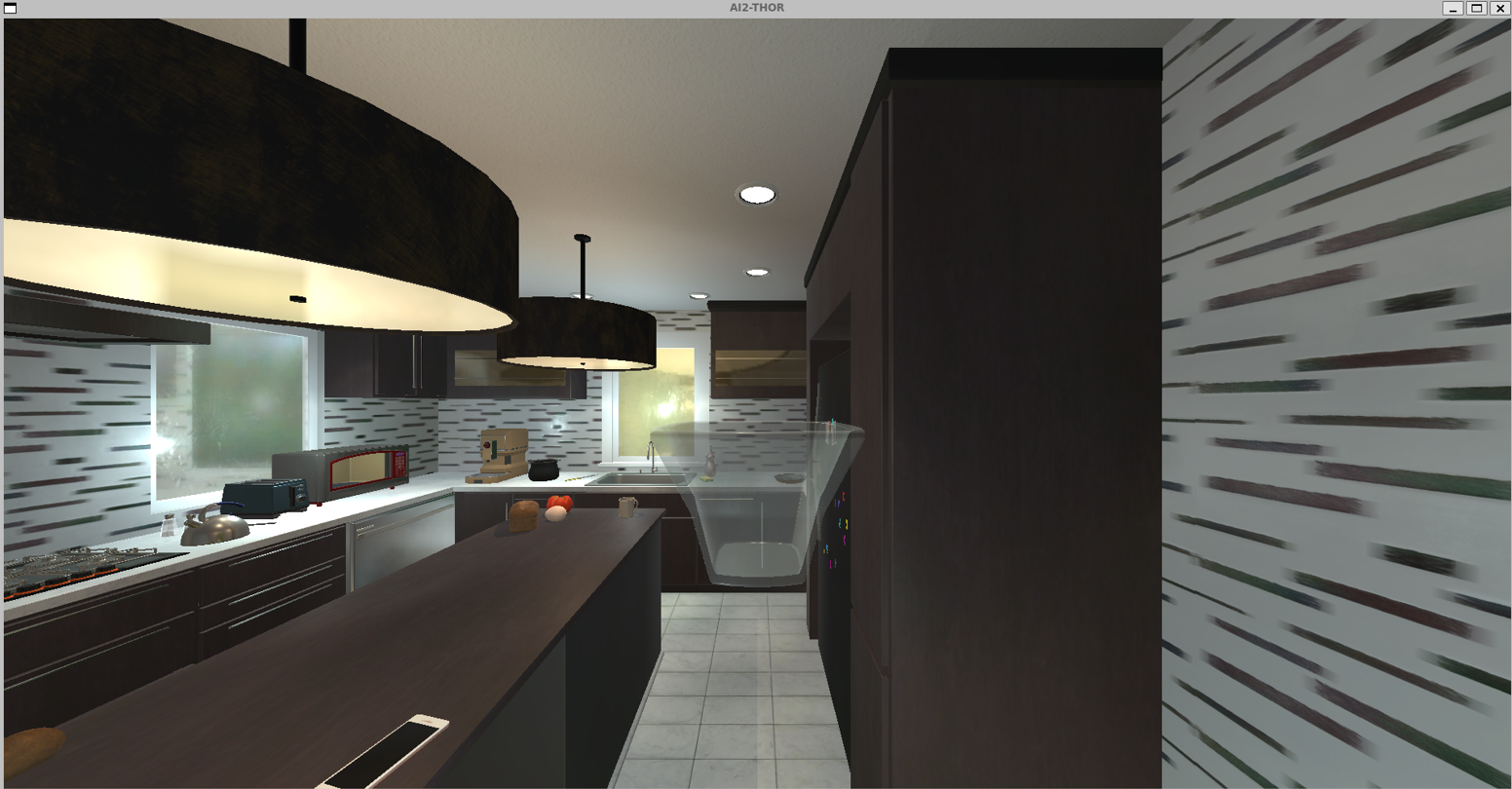}
    \caption{A container in FloorPlan2}
    \label{fig:transparent_container}
\end{subfigure}

\caption{Examples of various objects of different shape, size, color, mass, and opacity in different kitchen environments. These images showcase the diversity of objects the perception-enhanced agent encounters in its training, highlighting its ability to identify, navigate to, and interact with target objects efficiently.}
\label{fig:object_examples}
\end{figure*}

\section{Results}
\label{sec:results}

In this section, we present the results of our experiments, comparing the performance of the perception-enhanced agent with the baseline agent. We provide both quantitative metrics and qualitative analysis.

\subsection{Quantitative Analysis}

Table~\ref{tab:performance_comparison} summarizes the performance metrics for all agents.

\begin{table}[htbp]
\centering
\caption{Performance comparison of agents across all four environments (floor plans). Results are averaged over five runs with standard deviations.}
\label{tab:performance_comparison}
\begin{adjustbox}{max width=\linewidth}
\begin{tabular}{@{}lccc@{}}
\toprule
\textbf{Metric} & \textbf{Perception-Enhanced} & \textbf{Baseline} \\
\midrule
Success Rate (\%) & $73.5 \pm 2.1$ & $48.2 \pm 4.5$ \\
Avg. Cumulative Reward & $136.4 \pm 5.6$ & $81.1 \pm 9.3$ \\
Navigation Efficiency (\%) & $82.1 \pm 1.9$  & $61.7 \pm 3.1$ \\
Interaction Efficiency & $1.2 \pm 0.1$ & $2.1 \pm 0.3$ \\
\bottomrule
\end{tabular}
\end{adjustbox}
\end{table}

\subsubsection{Success Rate}

The perception-enhanced agent achieves a significantly higher success rate compared to the baseline, with an improvement of $52.5\%$. This indicates the agent's enhanced ability to locate and interact with target objects effectively.

\subsubsection{Average Cumulative Reward}

Our agent attains a $68.2\%$ increase in average cumulative reward over the baseline, reflecting more efficient policies and better adherence to the designed reward function.

\subsubsection{Navigation Efficiency}

An increase of $33.1\%$ in navigation efficiency demonstrates that the perception-enhanced agent follows paths closer to the optimal route, reducing unnecessary movements.

\subsubsection{Interaction Efficiency}

The agent requires fewer interaction attempts to successfully complete tasks, indicating better decision-making regarding when and how to interact with objects.

\subsection{Qualitative Analysis}

As shown in Figure~\ref{fig:object_examples}, the perception-enhanced agent effectively identifies and interacts with target objects with varied properties like shape, size, color, mass, opacity and so on, demonstrating versatility in handling different objectives in diverse environments. This capability highlights the robustness of the integrated perception models (SAM and YOLOv5), enabling the agent to navigate efficiently and interact precisely, regardless of the object's physical attributes or spatial configuration. The agent's adaptability across all four environments reinforces the effectiveness of combining advanced perception with reinforcement learning for complex interaction tasks.

\section{Discussion}
\label{sec:discussion}

Our results demonstrate that integrating vision foundation models with reinforcement learning agents can significantly enhance performance in object interaction tasks. The perception-enhanced agent outperforms baseline agent across all metrics, indicating that advanced perception contributes to more effective decision-making.

\subsection{Impact of Advanced Perception}

The use of SAM and YOLOv5 provides the agent with detailed semantic and spatial information, enabling it to better understand the environment. This enriched perception allows the agent to:

\begin{itemize}
    \item \textbf{Disambiguate Objects}: Correctly identify target objects among similar items.
    \item \textbf{Plan Efficient Paths}: Utilize spatial awareness to navigate around obstacles.
    \item \textbf{Optimize Interactions}: Position itself optimally for interaction tasks, reducing failed attempts.
\end{itemize}

\subsection{Policy Learning}

The agent's improved performance suggests that the policy network effectively leverages the enhanced perception. The feature representation $\phi(s_t)$ captures essential information without overwhelming the network, highlighting the importance of careful feature design.

\subsection{Computational Considerations}

Although the integration of large models increases computational demands, our optimizations mitigate the impact. The acceptable frame rate achieved suggests feasibility for real-time applications.

\subsection{Limitations}

Our experiments focused on specific environments and tasks. The agent's ability to generalize to new settings or handle unexpected changes remains an area for exploration. Incorporating mechanisms for continual learning or domain adaptation could enhance robustness.

\section{Conclusion}
\label{sec:conclusion}

We have presented a novel approach that integrates vision foundation models with reinforcement learning to enhance object interaction capabilities in simulated environments. By combining SAM and YOLOv5 with a PPO agent in the AI2-THOR environment, we demonstrate significant improvements over baseline agent in success rate, efficiency, and overall performance.

Our work highlights the potential of leveraging advanced perception models to enhance RL agents, enabling more effective interactions with complex environments. The positive results highlight the value of integrating state-of-the-art vision models into robotics applications.

We believe that our approach opens up new possibilities for developing sophisticated autonomous agents capable of operating in diverse and challenging settings.

\section{Future Work}
\label{sec:future_work}

Building on our findings, future research can focus on enhancing the agent's generalization to unseen environments via domain randomization or transfer learning, reducing computational overhead through model optimization techniques, and exploring real-world deployment by bridging the sim-to-real gap. Additionally, extending the agent's capabilities to handle multi-task learning and incorporating natural language instructions could further enhance its versatility.

\bibliographystyle{IEEEtran}
\bibliography{root}

\end{document}